\documentclass[fleqn,10pt]{wlscirep}
\usepackage{lineno,hyperref}

\usepackage[utf8]{inputenc}
\usepackage[T1]{fontenc}
\usepackage{multirow}
\usepackage{subfigure}
\usepackage{minted}
\usepackage{arydshln}
\usepackage{longtable}
\usepackage{subcaption}
\usepackage{graphicx}
\usepackage{booktabs,longtable}

\title{Evaluating Large Language Models for Stance Detection on Financial Targets from SEC Filing Reports and Earnings Call Transcripts}

\author[1]{Nikesh Gyawali}
\author[1,*]{Doina Caragea}
\author[2]{Alex Vasenkov}
\author[3]{Cornelia Caragea}
\affil[1]{Kansas State University, Department of Computer Science,  Manhattan, KS, 66506, USA}
\affil[2]{Mathinvestments, Inc., Huntsville, AL, 35801, USA}
\affil[3]{University of Illinois Chicago, Department of Computer Science, Chicago, IL, 60607, USA}
\affil[*]{dcaragea@ksu.edu}


\begin{abstract}
Financial narratives from U.S. Securities and Exchange Commission (SEC) filing reports and quarterly earnings call transcripts (ECTs) are very important for investors, auditors, and regulators. However, their length, financial jargon, and nuanced language make fine-grained analysis difficult. Prior sentiment analysis in the financial domain required a large, expensive labeled dataset, making the sentence-level stance towards specific financial targets challenging. In this work, we introduce a sentence-level corpus for stance detection focused on three core financial metrics: debt, earnings per share (EPS), and sales. The sentences were extracted from Form 10-K annual reports and ECTs, and labeled for stance (positive, negative, neutral) using the advanced ChatGPT-o3-pro model under rigorous human validation. Using this corpus, we conduct a systematic evaluation of modern large language models (LLMs) using zero-shot, few-shot, and Chain-of-Thought (CoT) prompting strategies. Our results show that few-shot with CoT prompting performs best compared to supervised baselines, and LLMs' performance varies across the SEC and ECT datasets. Our findings highlight the practical viability of leveraging LLMs for target-specific stance in the financial domain without requiring extensive labeled data. 
\end{abstract}
\begin{document}

\flushbottom
\maketitle
\thispagestyle{empty}

%\noindent Please note: Abbreviations should be introduced at the first mention in the main text – no abbreviations lists. Suggested structure of main text (not enforced) is provided below.

\section*{Introduction}

Financial narratives from U.S. Securities and Exchange Commission (SEC) filing reports and quarterly earnings call transcripts (ECT) constitute the most fundamental sources of information for a wide array of stakeholders. These documents provide comprehensive and fine-grained accounts of a company's financial health, operational performance, strategic initiatives, potential risks, past performance, and management's outlook on future prospects~\cite{bozanic2017sec,hassan2024economic}. They are highly valuable to investors, auditors, and regulators as they offer critical information about management’s perspective on key financial metrics such as debt, earnings per share (EPS), and sales. Despite their importance, these documents are often lengthy and complex, with convoluted sentence structures and use of specialized financial and legal terminologies, making manual analysis both time-consuming and labor-intensive.

Early work on sentiment analysis on financial texts relied on lexicon-based approaches, such as a lexicon dictionary~\cite{loughran2011liability} and traditional machine learning techniques~\cite{kogan2009predicting,koukaras2022stock, schumaker2009textual,chiong2018sentiment, antweiler2004all, koukaras2022stock} using feature representation like bag-of-words or Term Frequency-Inverse Document Frequency (TF-IDF)~\cite{kogan2009predicting,soong2021sentiment}. While simple, such techniques struggled with the domain-specific language and contextual nuances of financial texts~\cite{loughran2011liability, kearney2014textual}. Moreover, they classify whether the language is optimistic (positive sentiment) or pessimistic (negative sentiment), but often fail to capture how sentiment varies across different targets within the text. For example, an increase in debt might be seen as either a strategic opportunity or a major risk, depending on the context. 

While deep learning techniques like Recurrent Neural Networks (RNNs), Long Short-Term Memory (LSTM), and Gated Recurrent Unit (GRU) networks improved by modeling sequential and contextual information~\cite{kraus2017decision,sohangir2018big,mamillapalli2024gruvader}, the significant improvement came with the models like Bidirectional Encoder Representations from Transformers (BERT)~\cite{devlin2019bert,soong2021sentiment,karanikola2023financial} and its domain-specific adaptation on financial texts such as FinBERT~\cite{araci2019finbert, liu2021finbert}, which captures complex semantic relationships through large-scale pre-training. However, these models still struggle to provide a stance with respect to individual financial targets~\cite{liu2021finbert}. Additionally, they require thousands of labeled examples per class, which can be very expensive to gather and annotate.

The recent emergence of advanced Large Language Models (LLMs) provides a promising opportunity in the field of NLP and stance detection. With in-context learning, a single prompt can deliver near-state-of-the-art results on unseen stance datasets \cite{pangtey2025large,wang2025can}. The ability of these models to perform robustly with minimal or no task-specific labeled data provides significant importance for practical financial applications, given the high cost and effort required to create a large labeled dataset. While LLMs have been used in financial sentiment detection tasks, the research still lacks a systematic investigation on the precise task of detecting stances towards specific financial targets (debt, EPS, sales) at the sentence level. 

In this work, we construct a sentence-level corpus derived from ``Form 10-K (Annual Reports)'' filed with the U.S. Securities and Exchange Commission (SEC) and quarterly ECTs. The corpus consists of sentences that explicitly reference three financial targets--debt, EPS, and sales--or are relevant to these targets. We annotate these sentences with ChatGPT-o3-pro, an advanced reasoning model from OpenAI, with strict quality control with human validation. Utilizing this dataset, we perform a systematic comparison of various LLMs (Llama3.3, Gemma3, Mistral 3, and GPT-4.1-Mini) under zero-shot, few-shot, and chain-of-thought prompting scenarios. This approach is important for assessing the potential of LLMs in real-world financial analysis scenarios where the availability of extensive, target-specific annotated data is often limited. 

We make our prompts, data, and code publicly available for reproducibility and advanced research on LLM-based financial stance detection.

\section*{Related Works}
Research on sentiment analysis and stance detection in financial text has progressed from lexicon-based models and traditional machine learning models to transformers and large language models. Loughran and McDonald~\cite{loughran2011liability} showed that general negative word dictionaries misclassify common financial terms and proposed a domain-tailored lexicon that better captures positive/negative tone in 10-K reports, linking those sentiment measures to stock market reactions (e.g., returns and volatility). Mukherjee et al.~\cite{mukherjee2022ectsum} introduced \textsc{ECTSum}, a dataset curated by summarizing long ECTs. Various works utilized traditional machine learning techniques including regression models~\cite{kogan2009predicting,koukaras2022stock}, Support Vector Machines (SVMs)~\cite{schumaker2009textual,chiong2018sentiment}, Naive Bayes classifiers~\cite{antweiler2004all}, and tree-based methods like Random Forests~\cite{koukaras2022stock} that typically use feature representation like bag-of-words or Term Frequency-Inverse Document Frequency (TF-IDF)~\cite{kogan2009predicting,soong2021sentiment}. Similarly, deep learning techniques like Recurrent Neural Networks (RNNs), Long Short-Term Memory (LSTM), and Gated Recurrent Unit (GRU) networks have been used in sentiment analysis in the financial domain~\cite{kraus2017decision,sohangir2018big,mamillapalli2024gruvader}. 

More recent works utilize Bidirectional Encoder Representations from Transformers (BERT)~\cite{devlin2019bert} and its domain-specific adaptation on financial texts such as FinBERT~\cite{araci2019finbert, liu2021finbert, peng2021domain, karanikola2023financial, cicekyurt2025enhancing}. Singh et al.~\cite{singh2023fin} introduced Fin-STance, a deep learning-based multi-task model specifically designed for detecting both financial stance and sentiment from financial data.

The most recent LLMs trained with a massive volume of texts show superior natural language understanding and outperform smaller transformer-based models like FinBERT~\cite{kang2025comparative}. Various works have looked into utilizing the power of such LLMs in sentiment analysis and research in the financial domain~\cite{fatouros2023transforming, guo2023chatgpt, li2023chatgpt, zhang2023instruct, feng2025unleashing, wei2025large, huang2024open}. Wang and Brorsson~\cite{wang2025can} study various zero-shot, few-shot, and fine-tuning-based approaches with LLMs for financial text analysis. A comprehensive review of methods for sentiment (stance) analysis in financial texts is conducted by Du et al.~\cite{du2024financial,du2025natural}. In addition, Nie et al.~\cite{nie2024survey} provide an extensive survey of the applications of LLMs in the financial domain.

\begin{table*}[t]
\centering
\caption{Illustrative examples of instances from the ECT and SEC datasets. Sample instances are presented for each combination of financial target (debt, EPS, and sales) and stance class (Positive, Negative, or Neutral).}

\label{tab:ect_sec_examples}
\setlength{\tabcolsep}{4pt}
\renewcommand{\arraystretch}{1.1}
\begin{tabular}{@{}p{6.9cm} p{6.9cm} p{1.2cm} p{1.5cm}@{}}
\hline
\textbf{ECT} & \textbf{SEC} & \textbf{Target} & \textbf{Stance} \\
\hline
\textit{as a result of the strong EBITDA growth and free cash flow, we're on a path to see adjusted net leverage drop to approximately 3x by the end of the year, absent any other actions.} &
\textit{Total debt decreased \$6.4 million to \$194.4 million at December 31, 2020 from \$200.8 million at December 31, 2019.} &
debt & Positive \\
\hdashline
\textit{we estimate at least \$0.50 of adjusted eps accretion next year as the business returns toward pre-covid levels.} &
\textit{Weighted-average diluted shares outstanding in 2019 declined 2.8 percent year-on-year which benefited earnings per share.} &
EPS & Positive \\
\hdashline
\textit{the strong sales mix also led to excellent profitability.} &
\textit{In addition, selling prices increased year-on-year by 0.6 percent for full-year 2020, and lower raw-material costs reduced cost of sales as a percentage of sales.} &
sales & Positive \\
\hdashline
\textit{adjusted net leverage was 3.7 times at year-end, up slightly from Q3 reflecting the impact of the input-cost increases on EBITDA.} &
\textit{We cannot assure that our operating performance, cash flow and capital resources will be sufficient to repay our debt in the future.} &
debt & Negative \\
\hdashline
\textit{when combined, we estimate these two factors had more than a \$0.20 negative impact on adjusted eps in the quarter.} &
\textit{Divestiture impacts include the lost operating income from divested businesses, which decreased earnings per diluted share by 3 cents year-on-year for 2020.} &
EPS & Negative \\
\hdashline
\textit{so, we announced a 10\% decrease in the business for the segment, but that was really driven by two major factors.} &
\textit{Total sales decreased 14 percent in Mexico, which included decreased organic sales of 12 percent.} &
sales & Negative \\
\hdashline
\textit{net debt finished the year at just under \$1.2 billion.} &
\textit{As of December 31, 2019, we had approximately \$21.6 million of LIBOR-based debt.} &
debt & Neutral \\
\hdashline
\textit{on this slide, you can see the components that impacted our operating margins and earnings per share performance as compared to Q1 last year.} &
\textit{A discussion related to the components of year-on-year changes in operating-income margin and earnings per diluted share follows: Organic growth/productivity and other.} &
EPS & Neutral \\
\hdashline
\textit{just can you sustain that sales momentum?} &
\textit{Sales grew in home improvement and home care, while consumer health care and stationery and office declined.} &
sales & Neutral \\
\hline
\end{tabular}
\end{table*}

\section*{Financial Dataset}
% \subsection{SEC Form 10-K Annual Report}
The U.S. SEC Form 10-K is a comprehensive annual report mandated for publicly traded companies that provides a detailed overview of financial performance, operational activities, and corporate governance~\cite{cazier201610}. These reports are essential resources for investors, analysts, and regulatory bodies as they provide comprehensive insights into corporate strategic initiatives and identified risk factors. Specifically, ``\textit{Section 7, Management’s Discussion and Analysis (MD\&A)}'' serves as a primary focus for textual analysis, as it encompasses management’s interpretative narrative regarding financial outcomes, examination of recognized uncertainties, and forward-looking statements about prospective developments~\cite{amel2016information}.

% \subsection{Quarterly Earnings Call Transcripts}
Quarterly ECTs are a textual record of teleconferences held by a company's management with financial analysts, investors, and the media, usually following the release of quarterly financial results~\cite{mukherjee2022ectsum}. These transcripts serve as a valuable source of qualitative data by capturing the dialogues between senior executives and financial analysts and offer a more interactive and dynamic medium compared to static documents, such as SEC filings, allowing executives to explain performance metrics in greater depth, provide necessary context, and address analysts' queries directly~\cite{bushee2003open}.  

\begin{table*}[t]
\centering
\setlength{\tabcolsep}{5pt} % adjust column spacing
\caption{Class Label Distribution. Distribution of Positive, Negative, and Neutral stance labels for each financial target considered (debt, EPS, and sales) in SEC filing reports (SEC) and earnings call transcripts (ECT). Counts are shown separately for the training and test splits.}
\label{tab:data_stats}
\begin{tabular}{llcccccccc}
\hline
 & & \multicolumn{4}{c}{\textbf{Train}} & \multicolumn{4}{c}{\textbf{Test}} \\
\cline{3-6} \cline{7-10}
\textbf{Dataset} & \textbf{Target} & Positive & Negative & Neutral & Total & Positive & Negative & Neutral & Total \\
\hline
\multirow{3}{*}{SEC} 
 & debt  & 27  & 50  &  8 &  85 & 50  & 35  & 12 &  97 \\
 & EPS   & 10  & 32  &  3 &  45 & 14  & 14  &  9 &  37 \\
 & sales & 70  & 72  & 18 & 160 & 59  & 100 & 12 & 171 \\
\hline
\multirow{3}{*}{ECT} 
 & debt  & 73  &  7  & 10 &  90 & 65  & 12  & 20 &  97 \\
 & EPS   & 44  & 37  & 16 &  97 & 84  & 65  & 22 & 171 \\
 & sales & 79  & 52  & 14 & 145 & 129 & 127 & 39 & 295 \\
\hline
\end{tabular}
\end{table*}

% \subsection{Dataset Collection}
\label{sec:dataset_collection}

We utilize  SEC filing reports from two companies—MATIV Holdings Inc. and 3M Co.—using data from 2020-2021 as the training set and data from 2022-2024 as the test set. MATIV Holdings Inc. and 3M Co. operate in the same industry but have hugely different market caps. Similarly, we also use ECTs from these two companies from 2020-2021 as training data and 2022-2024 as a test. We transformed the text from SEC reports and ECTs into individual sentences using PyPDFLoader~\cite{pypdfloader} from the LangChain library. We then used the LLaMA-3 model to remove irrelevant sentences, keeping only those sentences containing key phrases related to financial targets of interest, including debt, EPS, and sales (see {SI Appendix A} for the prompt used for filtering relevant sentences). Our initial results indicated that LLaMa-3 generated a significant number of false positives, incorrectly classifying irrelevant sentences as relevant. To address this, we applied the same prompt to ChatGPT, whose superior reasoning abilities allowed us to filter out most of these false positives. We then annotated the relevant sentences using ChatGPT-o3-pro, an LLM with advanced reasoning, deep analytical thinking, and complex problem-solving capabilities~\cite{o3_openai_2025} (see  {SI Appendix B} for the prompt used to annotate the stance labels). The model generated both a stance label and a corresponding justification for each label. 

We validated ChatGPT-o3-pro's annotations and justifications of ECT sentences for MATIV Holdings Inc. by having human annotators evaluate their correctness. Annotators reviewed the stance label and accompanying justification produced by the model, indicating whether they agree with its reasoning. Across all evaluated targets—sales, EPS, and debt—human annotators showed over 97\% agreement with the justifications provided by ChatGPT-o3-pro, indicating a high level of alignment between the model's analytical output and human judgment.
Given this high agreement, we utilized ChatGPT annotations as ground truth for the ECTs of 3M Co. and SEC report statements from both companies (see Table~\ref{tab:ect_sec_examples} for sample instances for each target and stance class from two datasets). Table~\ref{tab:data_stats} shows the statistics of our final dataset. We have, on average,  122 training and 107 test instances per target in the SEC dataset, whereas the ECT dataset has, on average, 120 training and 193 test instances per target.

\section*{Methodology}
\subsection*{Large Language Models}
For a comprehensive comparative analysis, we employ four large language models: three open-source or open-weight models--Llama 3.3, Gemma3-27B, and Mistral 3 Small--and one proprietary model, ChatGPT 4.1-mini. A brief overview of each model is provided below:

% \begin{enumerate}
% \item \textbf{Meta LLaMa 3.3:}
\subsubsection*{Meta LLaMa 3.3}
The Llama 3.3 model (Llama3.3:70B) from Meta~\cite{meta_llama3p3_70b_2024} is a 70 billion-parameter instruction-tuned, text-only generative model specifically developed for instruction-following tasks. It is optimized for multilingual dialogue and uses an optimized transformer architecture~\cite{grattafiori2024llama}. The tuned version uses supervised fine-tuning (SFT) and reinforcement learning with human feedback (RLHF) to align with human preferences for helpfulness and safety. The model supports a context window of up to 128,000 tokens and is optimized for inference efficiency through Grouped-Query Attention. In evaluations, it outperforms both its predecessor (Llama 3.1-70B) and larger models like Llama 3.1-405B. Consequently, Llama 3.3 provides a robust and efficient option for research and production use in conversational AI.

\subsubsection*{Gemma-3-27B} Gemma 3 (Gemma3:27B) is a 27 billion-parameter instruction-tuned model from Google DeepMind~\cite{team2025gemma}. This model employs a multimodal architecture with vision understanding abilities, extends multilingual coverage, and has a longer context of at least 128,000 tokens. Using distillation-based pre-training and a novel post-training alignment phase, Gemma 3 significantly improves the math, chat, and instruction-following abilities, making its capabilities comparable to the more advanced and larger Gemini-1.5-pro model~\cite{team2025gemma}.

\subsubsection*{Mistral 3 Small} The Mistral Small 3 (Mistral3:24B) is a 24-billion-parameter, instruction-tuned decoder-only transformer model optimized for low latency token generation \cite{mistral2025small31} and has a context window of 32,000 tokens. The combination of strong multilingual coverage, competitive reasoning ability, and high throughput makes Mistral Small 3 a compelling open-weight foundation model for research and production-grade conversational systems that requires low latency and modest hardware~\cite{mistral2025small31}.

\subsubsection*{ChatGPT 4.1-mini}
The ChatGPT 4.1-mini is OpenAI's ``mini'' variant of the proprietary GPT-4.1 family~\cite{openai2025api}. We utilized the \textit{gpt-4.1-mini-2025-04-14} model snapshot for our performance comparison. This model has a very large context window of 1 million with full multimodal (text + image) input support. The GPT-4.1 mini is a fast and efficient small model, delivering significant improvements compared to the GPT-4o mini in instruction-following, coding, and overall intelligence~\cite{openai2025releaseNotes}.

 LLaMa 3.3, Gemma-3-27B and Mistral 3 Small models are instruction-tuned, transformer-based LLMs with sufficiently large context windows. GPT 4.1-mini, a proprietary model, has its exact parameter count undisclosed. OpenAI confirms it is considerably lighter than the full GPT-4.1 model~\cite{openai2025releaseNotes}. Because its scale is roughly comparable, we selected GPT 4.1-mini as a benchmark when comparing the two open-weight alternatives--Llama 3.3-70B.

\subsection*{Experimental Setup}
In our study, we experiment with LLMs on two primary datasets: the SEC reports and the ECTs. For both datasets, we study the usefulness of providing relevant background information about the company as {\it context} for the models in the prompt. Specifically, we extract ``\textit{Section 7 Management's Discussion and Analysis of Financial Condition}'' from the SEC report and the complete transcript from quarterly earnings calls, respectively. We experimented with three scenarios--zero-shot, few-shot, and context usage from the transcripts--each with and without chain-of-thought (CoT) reasoning prompts or demonstrations.

\noindent
\textbf{Context usage scenarios:} We evaluate the impact of providing background information about the company as additional context. We specifically investigate three scenarios:
\begin{enumerate}
    \item \textbf{No context:} In this scenario, neither Section 7 from the SEC report nor the ECTs are included in the prompt as additional context to the LLMs. This serves as a control for our experiment to study how adding additional context about the company affects the model's performance. 

    \item \textbf{Full context:} In this scenario, we provide the entire ``\emph{Section 7, Management’s Discussion and Analysis (MD\&A)}'' content from the SEC report or the complete ECT as additional context to the LLMs in their corresponding prompts for SEC and ECT, respectively.

    \item \textbf{Summarized context:} In this scenario, given that ``\emph{Section 7, Management’s Discussion and Analysis (MD\&A)}'' of the annual SEC reports spans over 20 pages on average, we summarize the Section 7 content from the SEC report using ChatGPT-o3-pro model to create more concise background information. Similarly, we also summarize the entire ECT using the ChatGPT-o3-pro model to study the differences in the performance of LLMs using summarized content.
    
\end{enumerate}

\noindent
\textbf{Prompting scenarios:} We evaluate  three prompting scenarios:
\begin{enumerate}
\item \textbf{Zero-shot setting:} In the zero-shot scenario, we evaluate how well LLMs perform without any labeled examples. We examine the performance across the three context conditions described above: (1) zero-shot with no context, (2) zero-shot with full context, and (3) zero-shot with the summarized context. This enables us to assess how the background information affects the model's ability to generalize without exposure to task-specific examples.
\item \textbf{Few-shot setting:} In the few-shot setting, we include a small number of labeled examples from the training set in the prompt to guide the LLM's predictions. We investigate two strategies for few-shot example selection: (1) random sampling and (2) selecting examples from the training set that are most semantically similar to the test instance. Similar examples are selected based on the highest cosine similarity between the test instance and the examples from the training set. The embeddings to calculate the cosine similarity are generated by using Sentence-BERT~\cite{reimers2019sentence}. We explore the effect of varying the number of examples, $k$, per stance class ($k=1,5,10$) for each target and dataset.  Similar to the zero-shot setting, we also test three background information configurations: (1) a few-shot with no context, (2) a few-shot with full context, and (3) a few-shot with summarized context.
\item \textbf{Chain-of-Thought setting:} In the Chain-of-Thought (CoT) setting, we prompt the LLMs to generate intermediate reasoning steps before arriving at a final prediction. This approach is particularly important for tasks that require multi-step reasoning. We evaluate the CoT performance using the same three context usage scenarios: (1) No context, (2) Full context, and (3) Summarized context. We also evaluate CoT performance in zero-shot and few-shot settings. This allows us to investigate how explicit reasoning steps combined with varying levels of contextual input and class-specific examples impact model performance and generalization.
\end{enumerate}

\begin{table}[t]
    \centering
    \setlength{\tabcolsep}{5pt} % adjust column spacing

    \caption{Few-shot classification accuracy (\%) on the ECT and SEC datasets. ``Random'' refers to using randomly selected few-shot examples, and ``Most similar'' refers to using examples from the training set that are most semantically similar to the test instance. Values are the mean $\pm$ standard deviation over three runs. For each model–dataset pair, the higher score between the two sampling strategies is \textbf{highlighted}.}

    \label{tab:random_vs_most_similar}
    \begin{tabular}{lcccc}
        \hline
        & \multicolumn{2}{c}{\textbf{ECT}} & \multicolumn{2}{c}{\textbf{SEC}} \\
        \cline{2-3} \cline{4-5}
        \textbf{Model} & Random & Most similar & Random & Most similar \\
        \hline
        GPT-4.1-Mini   & 86.73 $\pm$ 0.52 & \textbf{87.60 $\pm$ 0.75} & 83.65 $\pm$ 1.22 & \textbf{85.29 $\pm$ 1.20} \\
        Gemma3:27B     & 85.63 $\pm$ 0.53 & \textbf{86.43 $\pm$ 0.45} & 76.55 $\pm$ 0.87 & \textbf{79.21 $\pm$ 0.96} \\
        Llama3.3:70B   & 84.36 $\pm$ 1.76 & \textbf{87.04 $\pm$ 2.07} & 79.14 $\pm$ 2.10 & \textbf{79.42 $\pm$ 2.35} \\
        Mistral:24B    & 76.49 $\pm$ 2.10 & \textbf{80.11 $\pm$ 2.03} & 62.69 $\pm$ 3.01 & \textbf{67.07 $\pm$ 2.58} \\
        \hline
    \end{tabular}
\end{table}

\begin{figure*}[t]
  \centering

  \includegraphics[width=\linewidth]{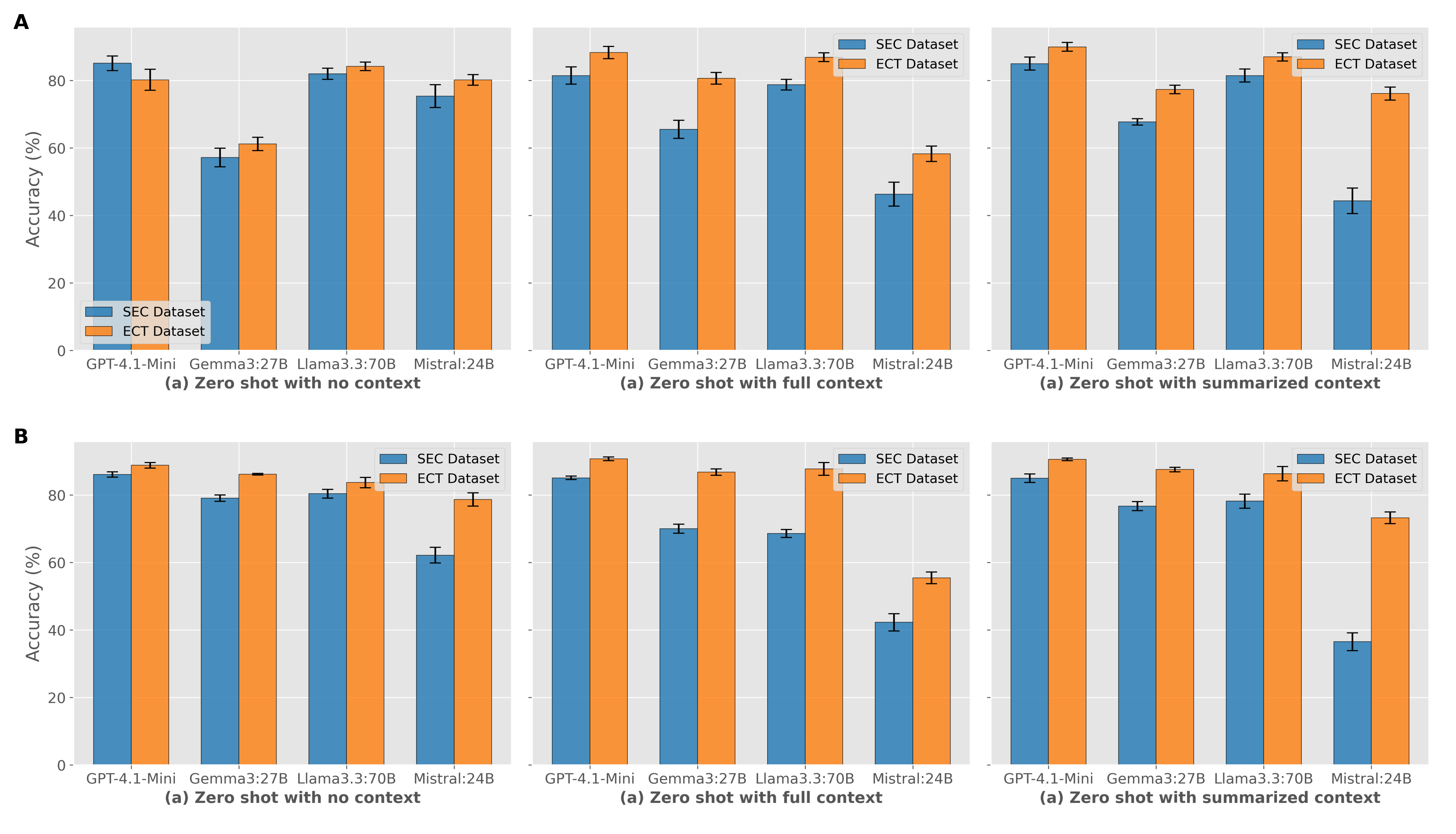}
   
  \caption{Zero-shot accuracy of models on the SEC and ECT datasets. The top row (Panel A) presents results with chain-of-thought (CoT) prompting, and the bottom row (Panel B) presents results without CoT prompting. Each condition is evaluated across three transcript-usage scenarios: (a) no transcript context, (b) full transcript context, and (c) summarized context.}
  \label{fig:zero-shot-cot-combined}
\end{figure*}

% \begin{figure}[!ht]
%   \centering
%   % --- top panel ------------------------------------------------------------
%   \begin{subfigure}{\textwidth}   % full width of the two-column page
%     \centering
%     \includegraphics[width=\textwidth]{figures/zero-shot-with_cot-transcript.png}
%     \caption{{\bf With chain-of-thought prompts.}}
%     \label{fig:zero-shot-with-cot-transcript}
%   \end{subfigure}

%   % \vspace{1.0em}   % vertical gap; trim or enlarge to taste

%   % --- bottom panel ---------------------------------------------------------
%   \begin{subfigure}{\textwidth}
%     \centering
%     \includegraphics[width=\textwidth]{figures/zero-shot-no_cot-transcript.png}
%     \caption{{\bf Without chain-of-thought prompts.}}
%     \label{fig:zero-shot-no_cot-transcript}
%   \end{subfigure}

%   % --- overall caption ------------------------------------------------------
%   \caption{\textbf{Zero-shot accuracy of models on the SEC and ECT datasets (A) \textit{with} and (B) \textit{without} chain-of-thought (CoT) prompting.} Each of these models and CoT settings is evaluated across three transcript-usage scenarios: (a) no transcript context,
%            (b) full transcript context, and (c) summarized context.}
%   \label{fig:zero-shot-cot-combined}
% \end{figure}

\section*{Results and Discussion}

The numeric results for both datasets, encompassing all models and experimental configurations, are summarized in the {SI Appendix E, Tables S1-S12}.

Our experimental results show that overall, ChatGPT-4.1-mini achieves the highest performance with an average accuracy across all experimental setups of 87.79\% $\pm$ 1.47, followed by Llama3.3:70B (83.02\% $\pm$ 1.98). Gemma3:24B shows comparable performance with an accuracy of 81.21\% $\pm$ 1.48 while Mistral:24B performs the worst with 68.6\% $\pm$ 2.71 accuracy. 
In what follows, we discuss the detailed findings from our experiments across various experimental setups.

\begin{figure*}[t]
  \centering
  \includegraphics[width=\linewidth]{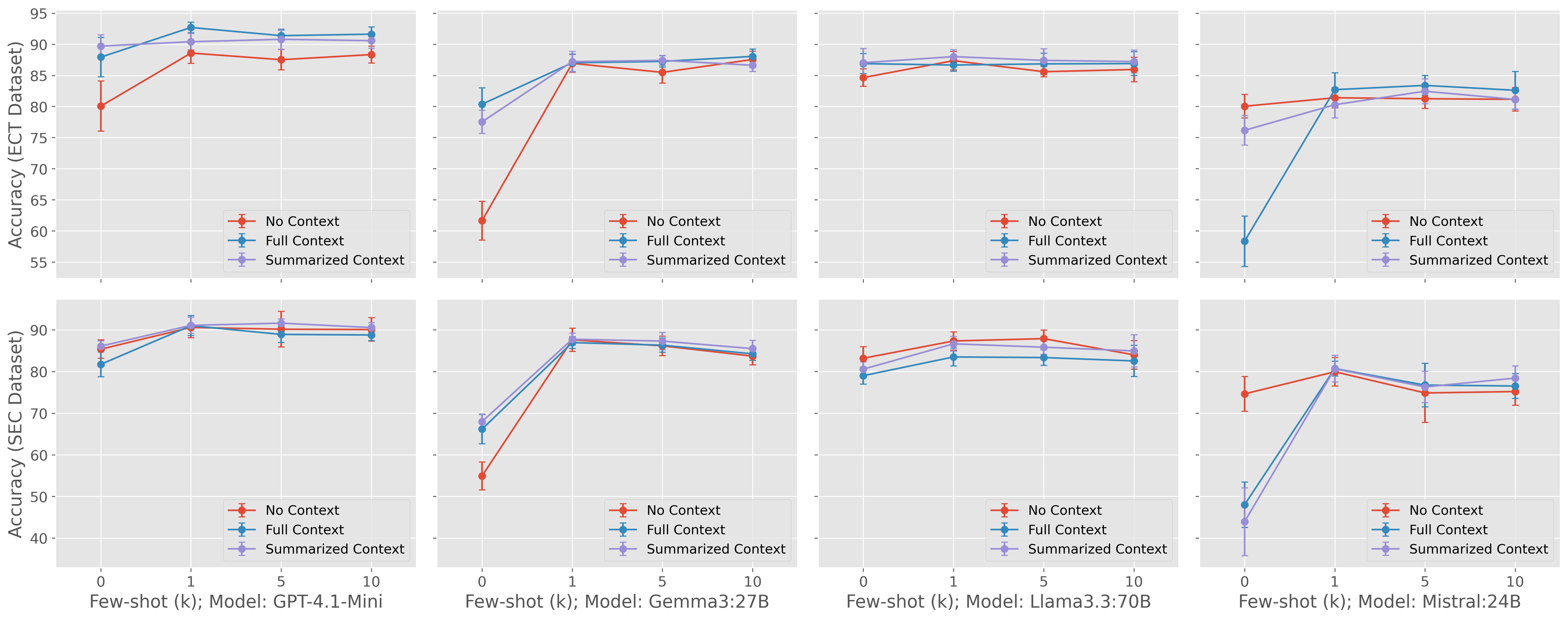}
  \caption{Context usage scenarios across different models on two datasets. 
  Few-shot classification accuracy is shown for the ECT dataset (top row) and the SEC dataset (bottom row) under three context-usage scenarios: (a) no context, (b) full context, and (c) summarized context, across four models. 
  Columns, from left to right, represent GPT-4.1-Mini, Gemma3-27B, Llama3-70B, and Mistral-24B. The variable $k$ indicates the number of most similar examples with a chain-of-thought demonstration per class.}
  \label{fig:context-usage-scenarios}
\end{figure*}

\subsection*{Context usage: No, Full, and Summarized}

Figure~\ref{fig:context-usage-scenarios} shows the few-shot performance of the models on the ECT dataset (top row) and the SEC dataset (bottom row) across three context usage scenarios. Within each figure, accuracy (mean $\pm$ std) is plotted against the number of few-shot examples $k=0, 1, 5, 10$ for three context usage scenarios:  No context (red), Full context (blue), and Summarized context (light purple). 
We find that incorporating contextual information--either full or summarized-- markedly improves the zero-shot performance on the ECT dataset for GPT-4.1-Mini and Gemma3:27B, whereas there is no significant improvement for the Llama3.3:70B model, and performance even degrades for the Mistral:24B model. For the SEC dataset, only Gemma3:27B model shows significant performance improvement using the context, while the performance of Mistral:24B shows a significant drop in the zero-shot setting using the context. For models that benefit from additional context, we find that models perform similarly in both full and summarized context usage, suggesting that the summarized context is nearly as effective as the full context.

As the number of few-shot examples increases, we find that all model performs similarly regardless of the context provided (no, full, or summarized context) for both datasets. In this scenario, contextual information provides minimal additional benefit, indicating that models prioritize in-context learning from the provided few-shot examples over utilizing the contextual information. This finding suggests that when a sufficient few-shot examples are available, extensive context becomes less important for achieving a good performance.

\subsection*{Random vs. Semantically Similar Few-shot Examples}

Table~\ref{tab:random_vs_most_similar} shows the performance comparison of various LLMs when using randomly selected examples versus semantically most similar few-shot examples across both ECT and SEC datasets. Across all models and both datasets, selecting the most similar examples consistently improves the accuracy over randomly selecting the few-shot examples. For the ECT dataset, the average performance improvement across all models when using semantically similar examples is 2\%, whereas for the SEC data, the average improvement is slightly higher, 2.24\%. The most notable performance improvement is observed with the Mistral:24B, with an average accuracy improvement of 4\% across both datasets. The remaining models show modest performance improvement with average improvement of 1.73\% for Gemma3:24B, 1.48\% for Llama3.3:70B, and 1.26\% for GPT-4.1-Mini.

Our experimental results show that selecting the semantically most similar few-shot examples consistently yields superior performance compared to selecting few-shot examples randomly. These findings indicate that carefully selecting examples that are semantically similar to the test instance provides more relevant context and clearer guidance for LLMs, thereby improving stance detection across diverse linguistic scenarios.

\begin{figure*}[!ht]
  \centering
  \includegraphics[width=\linewidth]{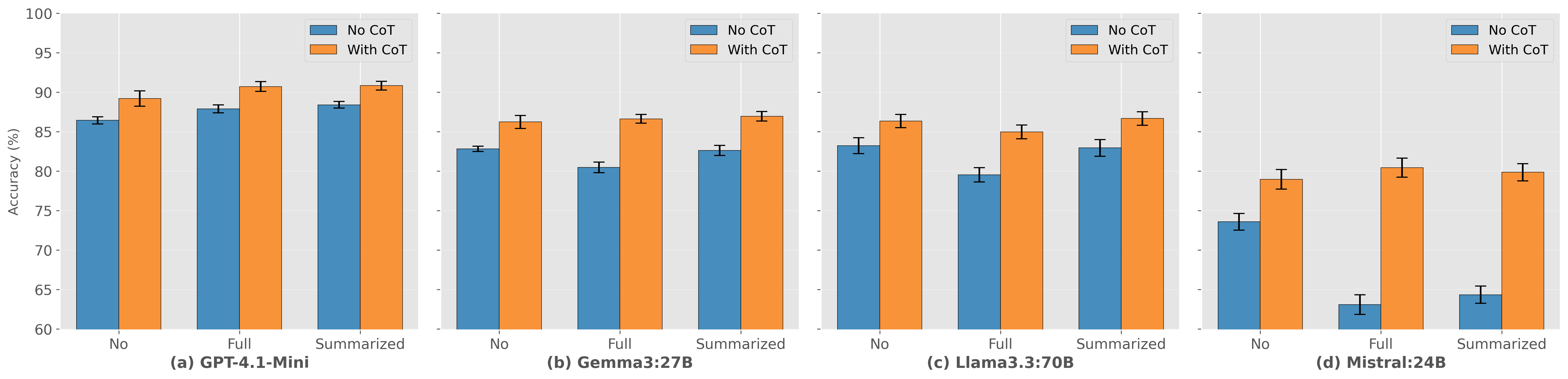}
  \caption{Few-shot performance of transcript usage scenarios \emph{with} and \emph{without} chain-of-thought (CoT) prompts. Accuracy is shown for four models--(a) GPT-4.1-mini, (b) LLaMA3.3:70B, (c) Gemma3:24B, and (d) Mistral:24B. The result is averaged for both SEC and ECT data.}
  \label{fig:cot-model-comparision}
\end{figure*}

\begin{figure*}[!ht]
  \centering
  \includegraphics[width=\linewidth]{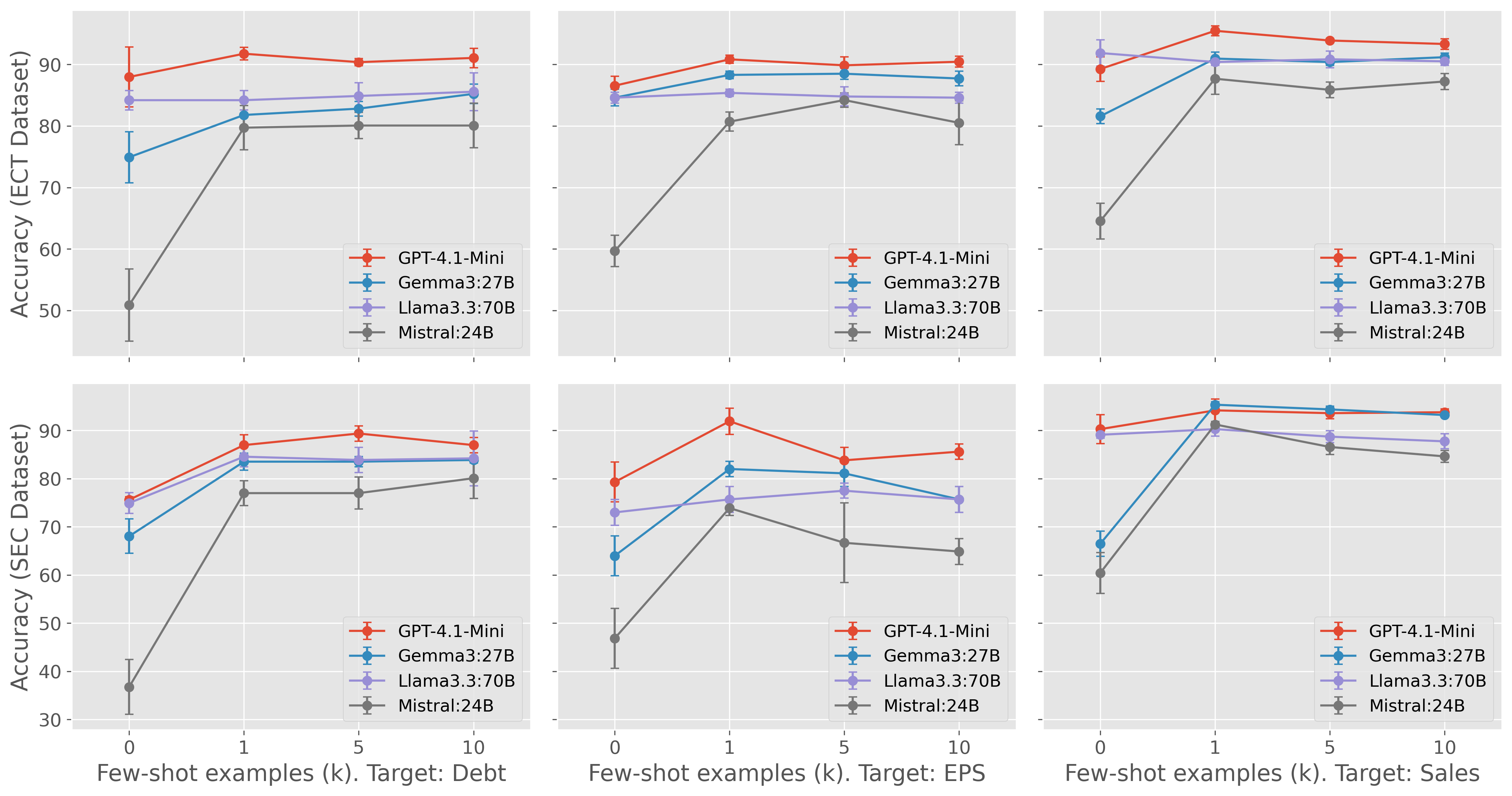}
  \caption{Few-shot with chain-of-thought accuracy on two datasets across various targets. Few-shot classification accuracy on the ECT dataset (top row) and the SEC dataset (bottom row) using chain-of-thought prompting for three targets—debt (left), EPS (centre), and sales (right). $k$ represents the number of most similar examples with a chain-of-thought demonstration per class. Error bars represent the standard deviation over three independent runs.}

  \label{fig:few-shot-comparison}
\end{figure*}

\subsection*{Effectiveness of Chain-of-Thought (CoT) Prompting}
Figure~\ref{fig:cot-model-comparision} shows the few-shot performance of various models, comparing scenarios with and without chain-of-thought (CoT) prompting demonstration across various transcript usage scenarios (see  {SI Appendix C} for an example). The results are averaged over both ECT and SEC datasets. We find that incorporating CoT reasoning in the few-shot examples consistently improved performance across all evaluated LLMs compared to the scenario without CoT demonstrations. The accuracy improvement with CoT demonstrations is 4.23\% $\pm$ 0.47 averaged across all models and transcript usage scenarios. The most significant improvement is observed for the Mistral:24B model, particularly when using Full and Summarized transcripts. Llama3.3:70B and Gemma3:27B show moderate improvement, while GPT-4.1-Mini shows the smallest improvement.

These findings suggest that CoT prompting explicitly encourages structured reasoning, enhancing the models' reasoning capabilities~\cite{wei2022chain}, which improves the performance in the stance detection task. This CoT prompting is especially beneficial for relatively smaller models, which may rely more on guided reasoning to bridge gaps in implicit understanding of text~\cite{ranaldi2024aligning}.

\subsection*{Few-shot Prompting vs. Zero-shot Prompting}
Figure~\ref{fig:few-shot-comparison} shows the performance of various LLMs under zero-shot and few-shot prompting strategies, using $k=1, 5, 10$ examples with CoT demonstration, across three targets--debt, eps, and sales--for both ECT and SEC datasets. We find that, on average, few-shot prompting with a single example $k=1$ consistently yields better performance over the zero-shot setting across all targets and datasets. The improvement is significant for the Mistral:24B model. All models except Llama3:70B show consistent improvement over zero-shot across all targets for $k=1, 5, 10$ examples in both datasets, suggesting the robustness of few-shot prompting in enhancing models' generalization and in-context learning~\cite{kojima2022large, brown2020language}. Interestingly, the most significant performance improvement is seen with $k=1$, implying that even a minimal example with CoT guidance can be effective. With $k=5, 10$ we do not see significant improvement, indicating that additional few-shot examples have diminishing returns as adding more examples increases the prompt length and increases the chance of noisy or conflicting chain-of-thoughts, thus decreasing accuracy~\cite{liu2024mind, levy2024same}. These findings highlight the general importance of few-shot learning for stance detection. They also show that model-specific variations exist, suggesting the importance of tuning the optimal number of examples,$k$, for different LLM architectures.

\subsection*{Performance across SEC and ECT Datasets}
Figure~\ref{fig:zero-shot-cot-combined} shows the performance of the LLMs on the SEC and ECT datasets under identical experimental conditions--specifically, with and without chain-of-thought prompts and varying usage of transcript contexts:(a) No context, (b) Full context, and (c) Summarized context. Similarly, Fig.~\ref{fig:few-shot-comparison} shows the performance of the LLMs on the ECT and SEC datasets using few-shot examples ($k=1, 5, 10$) with chain-of-thought (CoT) prompting across all three targets: debt, eps, and sales.

We find a consistent trend where models achieve similar higher accuracy on the ECT dataset as compared to the SEC dataset across all similar evaluated conditions. Similarly, across experiments with few-shot prompting with $k=1, 5, 10$ examples, all models consistently achieve higher accuracy on the ECT data than on the SEC data. This performance gap is attributed to fundamental differences in linguistic styles between these two datasets. ECT data consists of conversational statements and often narrates explicit financial changes (e.g., increases or decreases in debt, eps, or sales), providing clearer stance indicators. In contrast, the SEC dataset comprises formally structured sentences that are often numerically dense and embed key financial indicators in complex syntactic and semantic structures. Consequently, identifying the stance within the SEC dataset requires deeper quantitative reasoning capabilities, which continue to pose significant challenges for current LLMs (See {SI Appendix D} for further error analysis).

For example, in a SEC instance--``\textit{Our total debt to capital ratios, as calculated under the amended Credit Agreement, at December 31, 2022 and December 31, 2021 were 59.0\% and 65.1\%, respectively.}'', the stance is positive towards debt as the debt-to-capital ratio fell from 61.1\% to 59.0\%, indicating reduced leverage. However, identifying this requires the model to compute a relative change and understand its implications. Llama3.3:70B incorrectly classifies this as neutral, with justification--{\it The sentence provides a factual comparison of debt to capital ratios without expressing a clear positive or negative stance towards debt, merely stating the ratios for two years--which lacks understanding of the implication of this change in ratios}. In contrast, an ECT stance--``\textit{Net debt stands at \$13.3 billion, up approximately 2\% as we continue to invest in the business.}''--has a negative stance towards debt and presents a more direct cue (i.e., an explicit increase in debt), making it easier for the LLMs to correctly identify the negative stance. These examples highlight the greater complexity and reasoning involved in the SEC data, which contributes to the observed accuracy disparities.

\section*{Conclusion}
In this study, we systematically evaluated several modern LLMs for stance detection in financial texts at the sentence level, focusing on three financial metrics as targets--debt, EPS, and sales. We introduced a sentence-level stance detection corpus derived from SEC Form 10-K filings and quarterly earnings call transcripts. Using the advanced ChatGPT-o3-pro model, we annotated the sentences from SEC reports and earnings call transcripts with rigorous human validation. We evaluated multiple prompt-based learning strategies, including zero-shot, few-shot, and chain-of-thought (CoT) prompting. Our findings indicate that few-shot prompting enriched with CoT reasoning yields superior performance over zero-shot and without CoT prompting. Particularly, the GPT-4.1-Mini model consistently outperformed other evaluated LLMs, followed by the Llama3.3:70B model. Our analysis further highlights the importance of the selection of semantically similar few-shot examples in improving the model performance. Furthermore, we observed notable performance differences across two document types--ECT and SEC filing reports. While ECT data provided relatively easier stance detection due to its conversational nature and more explicit financial references, SEC reports posed greater challenges due to their formal complexity and numerical density, emphasizing the need for advanced contextual understanding and reasoning capabilities. Our findings highlight the practical viability of leveraging LLMs for the target-specific stance detection task in the financial domain without requiring extensive labeled data. 

\section*{Limitations}
Our study has a few limitations that should be acknowledged. First, the dataset is limited in scope as we exclusively focused on two companies--MATIV Holdings Inc. and 3M Co.--which may constrain the generalizability of the findings across broader financial contexts. Second, we rely on the ChatGPT-o3-pro model for dataset annotation. While the agreement with human validation is very high (over 97\%), the use of a language model for data annotation may introduce model-induced biases. Specifically, the results generated by ChatGPT-4.1-mini could be subject to such biases.

% \section*{Discussion}

% The Discussion should be succinct and must not contain subheadings.

% \section*{Methods}

% Topical subheadings are allowed. Authors must ensure that their Methods section includes adequate experimental and characterization data necessary for others in the field to reproduce their work.

% \noindent LaTeX formats citations and references automatically using the bibliography records in your .bib file, which you can edit via the project menu. Use the cite command for an inline citation, e.g.  \cite{Hao:gidmaps:2014}.

% For data citations of datasets uploaded to e.g. \emph{figshare}, please use the \verb|howpublished| option in the bib entry to specify the platform and the link, as in the \verb|Hao:gidmaps:2014| example in the sample bibliography file.

% Must include all authors, identified by initials, for example:
% A.A. conceived the experiment(s),  A.A. and B.A. conducted the experiment(s), C.A. and D.A. analysed the results.  All authors reviewed the results and approved the final version of the manuscript

% To include, in this order: \textbf{Accession codes} (where applicable); \textbf{Competing interests} (mandatory statement). 

% The corresponding author is responsible for submitting a \href{http://www.nature.com/srep/policies/index.html#competing}{competing interests statement} on behalf of all authors of the paper. This statement must be included in the submitted article file.

%\bibliographystyle{naturemag-doi}
%\bibliography{references}

\section*{Acknowledgments}
This research was supported by the National Science Foundation STTR Grant No. 2343777.

\section*{Author contributions}
A.V., C.C. and D.C. conceptualized the research. A.V., D.C and N.G. designed specific computational experiments. A.V. provided SEC Filing Reports and Earnings Call Transcripts data. N.G. processed and annotated the dataset. N.G. conducted all experiments. All authors reviewed the results.  N.G. drafted the first version of the manuscript.  D.C. supervised all aspects of the research and provided critical feedback. All authors read and approved the final manuscript.

\section*{Data availability}
The code and datasets generated during and/or analyzed during the current study are available in the GitHub repository \href{https://github.com/gnikesh/llm-financial-stance}{https://github.com/gnikesh/llm-financial-stance}.

\section*{Competing interests}
The authors declare no competing interests.

\section*{Additional information}

\subsection*{Supplementary Information (SI).} The manuscript  contains supplementary material, specifically Appendices A (LLM prompt for relevance filtering), B (LLM prompt for stance annotation), C (Chain-of-Thought example), D (Error Analysis), E (all experimental results - Tables S1-S12). 

\subsection*{Correspondence} Correspondence should be addressed to D.C.

% \section*{Supplementary Legends}
% \paragraph*{Section A}
% \label{supplementaryA}
% LLM prompt for relevance filtering

% \paragraph*{Section B}
% \label{supplementaryB}
% LLM prompt for stance annotation

% \paragraph*{Section C}
% \label{supplementaryC}
% Chain-of-Thought example

% \paragraph*{Section D}
% \label{supplementaryD}
% Error Analysis

% \paragraph*{Section E}
% \label{supplementaryE}
% All experiment results

\end{document}